\documentclass[10pt,twocolumn,letterpaper]{article}

\usepackage{cvpr}
\usepackage{times}
\usepackage{epsfig}
\usepackage{graphicx}
\usepackage{amsmath}
\usepackage{amssymb}

\usepackage{booktabs} % for \midrule in tables
\usepackage{float} % for figure and table [H] -> force HERE
\usepackage{booktabs}
\usepackage{csvsimple} %for csv tables
\usepackage{siunitx} % for roundnum
\usepackage{xspace} % for otehr command to handle space

\usepackage[pagebackref=true,breaklinks=true,colorlinks,bookmarks=false]{hyperref}
\cvprfinalcopy % *** Uncomment this line for the final submission

\def\cvprPaperID{1103} % *** Enter the CVPR Paper ID here

\begin{document}

%%%%%%%%% TITLE
\title{Leveraging Shape Completion for 3D Siamese Tracking}

\author{Silvio Giancola*, Jesus Zarzar*, and Bernard Ghanem\\
King Abdullah University of Science and Technology (KAUST), Saudi Arabia \\
{\tt\small \{silvio.giancola,jesusalejandro.zarzartorano,bernard.ghanem\}@kaust.edu.sa}\\
% \href{https://github.com/SilvioGiancola/ShapeCompletion3DTracking}{https://github.com/SilvioGiancola/ShapeCompletion3DTracking}
}

\maketitle

\newcommand\blfootnote[1]{%
  \begingroup
  \renewcommand\thefootnote{}\footnote{#1}%
  \addtocounter{footnote}{-1}%
  \endgroup
}
\blfootnote{* Both authors contributed equally to this work.}

\newcommand{\mysection}[1]{\vspace{2pt}\noindent\textbf{#1}}
\newcommand{\TODO}[1]{\textcolor{red}{\textbf{\textit{[TODO] #1}}}}
\newcommand{\BG}[1]{\textcolor{red}{\textbf{\textit{[BG] #1}}}}
\newcommand{\JZ}[1]{\textcolor{red}{\textbf{\textit{[JZ] #1}}}}
\newcommand{\SG}[1]{\textcolor{red}{\textbf{\textit{[SG] #1}}}}
\newcommand{\sota}{state-of-the-art\xspace}
\newcommand{\Sota}{State-of-the-art\xspace}
\newcommand{\Table}[1]{Table~\ref{tab:#1}}
\newcommand{\Figure}[1]{Figure~\ref{fig:#1}}
\newcommand{\Equation}[1]{Equation~\eqref{eq:#1}}
\newcommand{\Section}[1]{Section~\ref{sec:#1}}

%%%%%%%%% ABSTRACT
\vspace{-10pt}
\begin{abstract}
    Point clouds are challenging to process due to their sparsity, therefore autonomous vehicles rely more on appearance attributes than pure geometric features. However, 3D LIDAR perception can provide crucial information for urban navigation in challenging light or weather conditions. In this paper, we investigate the versatility of Shape Completion for 3D Object Tracking in LIDAR point clouds. We design a Siamese tracker that encodes model and candidate shapes into a compact latent representation. We regularize the encoding by enforcing the latent representation to decode into an object model shape. We observe that 3D object tracking and 3D shape completion complement each other. Learning a more meaningful latent representation shows better discriminatory capabilities, leading to improved tracking performance. We test our method on the KITTI Tracking set using car 3D bounding boxes. Our model reaches a \textbf{76.94\%} Success rate and \textbf{81.38\%} Precision for 3D Object Tracking, with the shape completion regularization leading to an improvement of \textbf{3\%} in both metrics.
\end{abstract}

%%%%%%%%% BODY TEXT

\vspace{-20pt}
\section{Introduction}

Autonomous driving is changing the way we envision human transportation. Introducing fully autonomous vehicles into our cities implies sharing the roads with existing vehicles.
Thus, it is imperative for autonomous vehicles to outperform humans in the task of driving.
Understanding the urban environment and the human driving process is crucial for an agent to become capable of achieving and exceeding human driving performance.
Accordingly, autonomous vehicles need to outperform human perception so to cope with an unbounded set of unpredictable situations.

An autonomous vehicle adapts its driving policy by understanding its environment.
Modules for Road Detection~\cite{dahlkamp2006self,caltagirone2018lidar} and Road-sign Recognition~\cite{greenhalgh2012real,timofte2014multi}  indicate to the car where and how to drive.
Object Detection methods~\cite{chen2017multi,qi2017frustum} constrain the vehicle's path in order to avoid collisions while Object Tracking algorithms~\cite{xiang2015learning,sharma2018beyond} predict their motion to anticipate danger.
Autonomous vehicles need to sense both appearance and geometric components of the environment to extrapolate the semantic information required for driving.
RGB cameras provide both appearance and geometric information by either inferring depth from single RGB cameras~\cite{Uhrig2017THREEDV,zhang2018deep} or by stereoscopy~\cite{yang2018segstereo,cheng2018learning}.
Depth and shape completion~\cite{ma2018self,jaritz2018sparse} are commonly used to improve the limited sensing capability of RGB sensors.

\begin{figure}[t]
    \centering
    \includegraphics[trim={4.5cm 2cm 4cm 2.5cm},clip,width=0.48\textwidth]{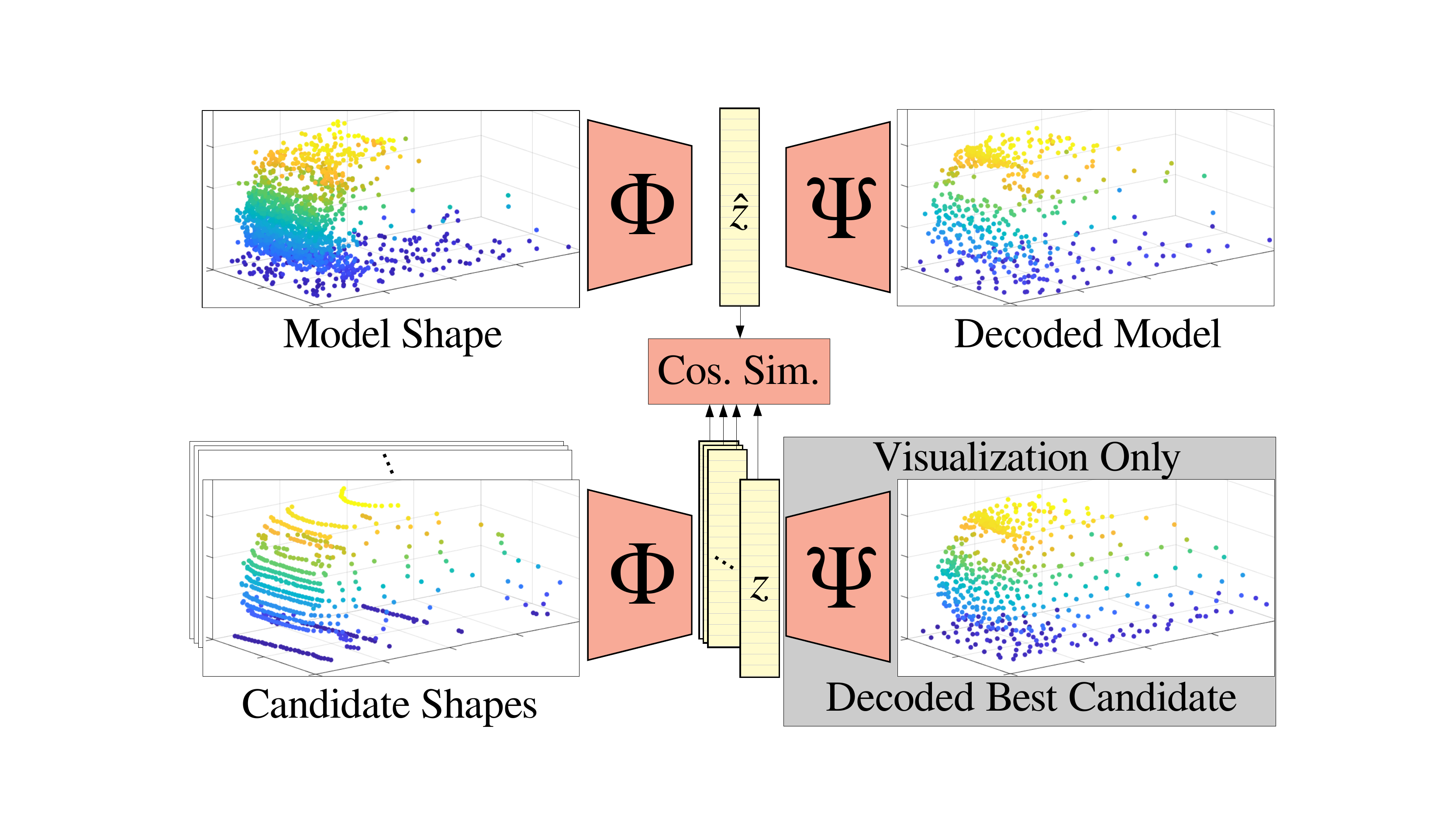}
    \caption{Our tracking model combines a Siamese network with an Auto-Encoder.
    The Siamese network encodes sparse 3D shapes into a latent representation $z$, in which shapes belonging to the same object have a high cosine similarity.
    By regularizing our tracker to auto-encode model shapes, we enforce that the encoder maps point clouds into meaningful representations.
    The effect of regularization is visualized by decoding a candidate shape.}
    \vspace{-10pt}
    \label{fig:Pooling}
\end{figure}

Alternatively, LIDAR systems directly sense geometry in a more accurate manner.
LIDAR sensors are less sensitive to light and weather conditions, so they provide more reliable information in a much larger range of driving conditions.
However, LIDARs generate sparse point clouds, not readily suitable for conventional CNN processing.
Most current works pre-process 3D point clouds for use in CNNs by either voxelizing the 3D space~\cite{ku2017joint,hua2018pointwise} or by projecting point clouds into a planar space~\cite{tatarchenko2018tangent,Kostrikov_2018_CVPR,chu2018surfconv,su2018splatnet}.
However, these methods lose fine-grained geometric details.
It is worthwhile to note that only a few works deal directly with point clouds~\cite{qi2017pointnet,achlioptas2018learning}.
We believe appearance information is insufficient to reach better-than-human driving performance, especially in challenging driving  environments.

In this work, we propose an online 3D Object Tracking method based purely on LIDAR.
First, we leverage geometric features computed from sparse point clouds using the shape-completion network proposed by Achlioptas~\etal~\cite{achlioptas2018learning}.
These features are used in a Siamese network to create a latent representation in which a cosine similarity matches partial object point clouds to a model shape.
Then, we regularize the encoding via an auto-encoder network to generate geometrically meaningful latent representations.
We expect improved tracking performance by enriching the latent representation with semantic geometric information from the given object.

Currently, the main challenges faced in tracking relate to
\textbf{(a)} similarity metrics,
\textbf{(b)} model updates, and
\textbf{(c)} occlusion handling.
Our 3D tracker tackles these three aspects by  
\textbf{(a)} using Siamese networks, which have been shown to achieve state-of-the-art performance on 2D visual object tracking, adapted for processing 3D LIDAR point clouds,
\textbf{(b)} leveraging the shape invariance in rigid bodies to generate a more complete model by aggregating its shape in time, and
\textbf{(c)} enforcing our model to understand shape regardless of occlusions through shape completion.

\mysection{Contributions:}
Our contributions are three-fold.
\textbf{(i)} To the best of our knowledge, we propose the first 3D Siamese tracker applied to point clouds rather than images.
\textbf{(ii)} We propose to regularize the Siamese network's latent space such that it resembles the latent space of a shape completion network.
\textbf{(iii)} We show that regularizing our network with semantic information results in better discrimination and tracking. To ensure reproducibility and to promote future research, 
all source code, trained model weights, and dataset results are publicly available\footnote{\href{https://github.com/SilvioGiancola/ShapeCompletion3DTracking}{https://github.com/SilvioGiancola/ShapeCompletion3DTracking}}.

\section{Related Work}

Our work takes insights from 
Object Tracking based on Siamese networks, 
Shape Representation and Completion based on Auto-encoders, and 
Search Strategy.

\mysection{Visual Object Tracking.}
Tracking is the task of identifying the trajectory of an object through time, either in images~\cite{vot_tpami,mot16} or in 3D space~\cite{luber2011people,song2013tracking}.
Visual tracking focuses on image \emph{patches} across consecutive frames, that represent visual attributes~\cite{vot_tpami}, objects~\cite{Muller_2018_ECCV}, people~\cite{luber2011people} or vehicles~\cite{geiger2013vision}.
The problem is commonly tackled by \emph{tracking-by-detection}, where a \emph{model representation} is built after the first frame and a \emph{search space} is constructed to trade off computational costs and dense space sampling.
Early works on tracking were based on Correlation Filtering~\cite{bertinetto2016staple}, but current better performing methods rely on deep CNNs~\cite{Jung_2018_ECCV} and Siamese networks~\cite{bromley1994signature}.
Bertinetto~\etal~\cite{bertinetto2016fully} introduced Siamese networks for visual object tracking.
They proposed a fully-convolutional Siamese network and achieved \sota performance on the VOT benchmark~\cite{vot_tpami}.
Recent Siamese trackers estimate boundary flows~\cite{Lei_2018_CVPR}, use contextual structure~\cite{He_2018_CVPR}, attention~\cite{Wang_2018_CVPR}, distraction~\cite{Zhu_2018_ECCV}, semantic information~\cite{Zhang_2018_ECCV}, triplet losses~\cite{Dong_2018_ECCV} and region proposal networks~\cite{Li_2018_CVPR} to improve tracking performance.
To the best of our knowledge, our work is the first 3D adaptation of Siamese networks for 3D point cloud tracking.

\mysection{3D Object Tracking.} 
3D Object Tracking tackles tracking from a geometric perspective. 
Instead of following appearance attributes using 2D bounding boxes (BBs), it computes the position of targets in the 3D world using geometry contained in 3D BBs. 3D object tracking either focuses on RGB-D  information~\cite{song2013tracking}, by mimicking the  2D object tracking methods but with an additional depth channel~\cite{Bibi_2016_CVPR,liu2018context}, or it focuses on purely geometric features~\cite{spinello2010layered,luber2011people}.
Recent work tackles 3D tracking using Bird Eye Views (BEV) of LIDAR point clouds~\cite{luo2018fast,yang2018pixor}.
Luo~\etal~\cite{luo2018fast} input multiple BEV frames to a deep CNN to perform detection, tracking, and motion forecasting.
Yang~\etal~\cite{yang2018pixor} used up to 35-channel BEV frames.
However, these methods lose fine-grained shape information by projecting the point cloud in the BEV.
LIDARs sense the environment from a single point of view inducing self-occlusion, \ie \emph{incomplete} observations~\cite{deng2017amodal}. Note that on images, occlusion leads to \emph{noisy} observations. %, hence amodal detection is prone to error~\cite{deng2017amodal}. 
Moreover, tracking assumes a BB prior for the first frame and, since the object is rigid, its extent in 3D space in successive frames remains constant.

\mysection{Shape Representation.}
3D shapes are complex entities to manage as they are usually sparse and lying in a continuous space, unlike images that are stored in dense and discrete matrices.
Several works focus on finding efficient geometric representations~\cite{tangelder2004survey} such as occupancy grids and TSDF cubes.
They are commonly used for 3D reconstruction~\cite{newcombe2011kinectfusion,Giancola_2018_CVPR_Workshops} but suffer from large-scale memory inefficiency and require a space discretization which loses fine-grained details.
Recent works compress 3D shapes using auto-encoders to efficiently handle geometric information~\cite{wu2015shapeNets,umetani2017quadMesh,dai2017shape}.
They typically encode-decode shapes into different representations.
Those auto-encoders provide a compact latent shape representation of down to 10 dimensions.
Alternatively, Kundu~\etal~\cite{kundu20183d} used RGB information to decode dense 3D meshes of vehicles using Fast RCNN~\cite{ren2015faster} and a differentiable Render-and-Compare loss.
Achlioptas~\etal~\cite{achlioptas2018learning} proposed to solve shape completion using an efficient auto-encoder based on PointNet~\cite{qi2017pointnet} for point cloud to point cloud auto-encoding. 
They regress partial point clouds into full shapes.
Alternatively, Stutz~\etal~\cite{Stutz2018VoxelCompletion} proposed an occupancy grid shape completion network based on a two-stage training process. 
Also, Engelmann~\etal~\cite{engelmann2016joint} proposed an energy minimization method that aligns shape and pose concurrently in stereo images.

\mysection{Search Strategy.}
Search spaces used in visual object tracking are generally dense (exhaustive).
Bertinetto~\etal~\cite{bertinetto2016fully} used correlation filtering methods to obtain a similarity score for the whole search space.
However, exhaustive search space strategies are not realistically transferable to the continuous and denser 3D space.
This is commonly solved by relying on Kalman filters, Particle filters, or Gaussian mixture models to reduce the search space by providing candidate object proposals~\cite{ristic2003beyond,zhang2015structural}.
At each frame, particles are sampled according to a probability distribution. 
Only the selected particles are observed and the probability distribution is updated according to the observation.
Recently, Karkus~\etal~\cite{karkus2018particle} proposed a learnable particle filter network.
In our experiments, we choose to disentangle the search space and the similarity function, a common practice done in 2D tracking, by using an approximation of the exhaustive search detailed in the experiments.

\section{Methodology}

Herein, we propose a 3D Siamese tracker with a regularization on its latent space.
The tracker is regularized to learn an encoding containing semantically meaningful information.
An overview of our network is shown in \Figure{Pooling}.

\subsection{Siamese Tracker}
Our 3D Siamese tracker takes as input a sequence of point clouds (tracklet), in which a given object exists, along with an initial 3D BB corresponding to the position of the object in the first frame.
For a frame at time $t$, a set of candidate shapes $\{\mathbf{x}^t_c\}$ are encoded into latent vectors $\{\mathbf{z}^t_c\}$ and compared with the latent vector $\mathbf{\hat{z}}^t$ from a model shape $\mathbf{\hat{x}}^t$.
The best candidate is selected to be the object in the current frame, and the model shape $\mathbf{\hat{x}}$ is updated accordingly.

\begin{figure}[htb]
    \centering
    \includegraphics[trim={0cm 12cm 2.8cm 0cm},clip,width=0.47\textwidth]{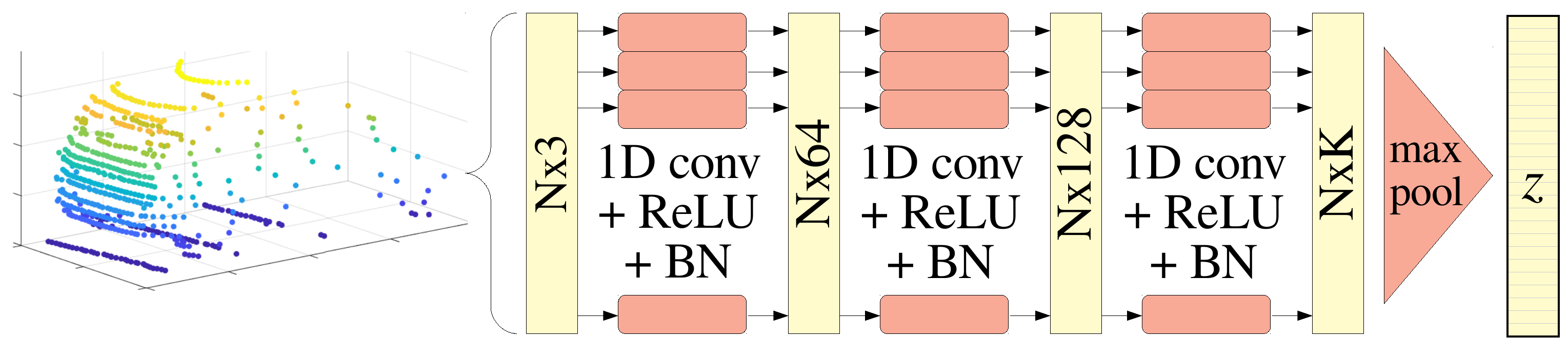}
    \caption{Our encoder takes as input a point cloud with $N=2048$ points. 
    Point clouds are encoded into a $K$-dimensional ($K=128$) latent vector $\mathbf{z}$ using 3 layers of 1D CNN with ReLU and BN.
    }
    \vspace{-15pt}
    \label{fig:Encoder}
\end{figure}

\mysection{Encoding.}
Our encoder $\Phi(\cdot)$ is inspired from previous work on shape completion by Achlioptas~\etal~\cite{achlioptas2018learning}.
This encoder consists of 3 layers of 1D-convolutions followed by ReLU layers~\cite{nair2010rectified} and BN layers~\cite{ioffe2015batch} with filter size [64, 128, K], as shown in \Figure{Encoder}.
The output of the last BN layer is followed by a max pooling across the points to obtain a K-dimensional latent vector.
We found $K=128$ to be a suitable size for the latent vector, as it provides the best trade-off between computational efficiency, latent space compactness, and tracking performance.
The input to our network is pre-processed to have $N=2048$ points by randomly discarding or duplicating points, so to use mini-batches in training.
Note that more than $96\%$ of the vehicles in the KITTI dataset have less than $2048$ points.
% We set $N=2048$ as this is a large enough number of points to maintain a good representation of an object's shape.
% We set $N=2048$, as it maintains a good representation of an object's shape and less than $4\%$ of the vehicles contains more point on KITTI.
% than $96\%$ of the vehicles in the KITTI dataset contains less points.
As compared to the network of \cite{achlioptas2018learning}, we leverage a more compact yet efficient latent space and a shallower network to reduce the size of the overall model from $\sim140$K to $\sim25$K parameters.

\mysection{Similarity Metric.}
% The encoder $\Phi(\cdot)$ calculates a latent representation from a point cloud.
The encoder $\Phi(\mathbf{x})$ extract a latent representation $\mathbf{z}$ from a point cloud $\mathbf{x}$.
To compare a pair of shapes $\mathbf{x}$ and $\mathbf{\hat{x}}$, we measure the cosine similarity between their respective latent vectors $\mathbf{z}$ and $\mathbf{\hat{z}}$ as per \Equation{CosineSimilarity}.

\begin{equation}
    CosSim(\mathbf{z},\mathbf{\hat{z}}) = \frac{\mathbf{z}^{\top} \mathbf{\hat{z}}  }{ \|\mathbf{z}\|_2 \|\mathbf{\hat{z}}\|_2 }
    \label{eq:CosineSimilarity}
\end{equation}

\mysection{Tracking Loss.}
For any given frame being used in training, we designate $\mathbf{x}$ to be the tracked object's point cloud and $\mathbf{\hat{x}}$ to be the ground truth model obtained by concatenating the object's point clouds across all frames in the tracklet.
We train our Siamese network to regress a function of the distance between a candidate shape $\mathbf{x}$ and the model shape $\mathbf{\hat{x}}$, according to \Equation{TrackingLoss}.
The poses of $\mathbf{x}$ and $\mathbf{\hat{x}}$ are parameterized by the 3 degrees of freedom of an object on a plane ($t_x$,$t_y$,$\alpha$).
The distance $d(\cdot,\cdot)$ is taken to be the L2-norm $\|\cdot\|_2$ of the difference between the parameterized poses.
The angle $\alpha$, given in degrees, is weighted with a factor of $\frac{1}{5}$ to have the same scale as $t_x$ and $t_y$ which are given in meters.
We chose the differentiable function $\rho(\cdot)$ to be a Gaussian with $\mu=0$, $\sigma=1$.
The purpose of $\rho(\cdot)$ is to soften the distance between positive and negative samples.
$\rho(\cdot)$ takes a value of one when the distance is zero and decays as the distance increases.
We then regress our similarity metric $CosSim(\cdot,\cdot)$ using an MSE loss as shown in \Equation{TrackingLoss}.
% Minimizing this loss for the network parameters encourages our encoder to encode partial and complete model shapes belonging to the same object into a latent space, in which they have a high cosine similarity.
Minimizing this loss encourages our encoder to increase the similarity between partial and complete shape to the same.
% similar object with different degree of completeness.

% closer representation of the same object.
% encode partial and complete model shapes belonging to the same object into a latent space, in which they have a high cosine similarity.

\vspace{-5pt}
\begin{equation}
    \label{eq:TrackingLoss}
    \mathcal{L}_{tr} =  \frac{1}{n} \sum_\mathbf{x} \Big( CosSim \big(\phi(\mathbf{x}), \phi(\mathbf{\hat{x}}) \big) -  \rho \big( d \left( \mathbf{x} , \mathbf{\hat{x}} \right) \big) \Big) ^2 
% \vspace{-5pt}
\end{equation}

\subsection{Shape Completion Regularization}

It is important to regularize the Siamese network in order to embed into the latent representation generative properties of shape that are useful in discrimination.
Such an embedding helps in generalizing to cases which aren't seen in training.
Our regularization enforces the Siamese network's latent space to lie within a shape representation space.
Such representation space embeds valuable semantic characteristics defining the object to track in a compact, meaningful, and efficient representation.
We provide qualitative evidence that the representation space learned by our model holds the required semantic characteristics by decoding latent representations as shown in \Figure{ModelCompletion}
Quantitative evidence is given through the improved tracking performances obtained in \Table{Ablation}.

\mysection{Decoding.}
Our decoder $\Psi(\mathbf{z})$ is inspired by the shape completion network employed by Achlioptas~\etal~\cite{achlioptas2018learning}.
Our decoder is composed of two fully connected layers that decode a $K=128$-dimensional latent vector $\mathbf{z}=\Phi(\mathbf{x})$ into $M$x$3$ values representing $M$ 3D points for a reconstructed shape $\mathbf{\tilde{x}}=\Psi(\Phi(\mathbf{x}))$.
We use $M=2048$ and a hidden layer of size $1024$ for a total of $\sim6.4$M parameters.
Alternatively, Achlioptas~\etal~\cite{achlioptas2018learning} decoded into a denser shape of $4096$ points, which requires more than twice the number of parameters in our decoder network.

\mysection{Completion Loss.}
Adding a completion loss as a regularizer for our Siamese network boosts the network's performance by enforcing the latent representation to hold semantic information of the tracked class.
While other works use the Earth Mover's Distance~\cite{rubner2000earth} to compare the model shape $\mathbf{\hat{x}}$ and the decoded model shape $\mathbf{\tilde{x}} = \Psi(\Phi(\mathbf{\hat{x}}))$, we use the Chamfer distance~\cite{fan2017point} (according to \Equation{CompletionLoss}), since it is simpler to compute~\cite{achlioptas2018learning}.
The tracking loss enforces encoded partial shapes to be similar to their respective encoded model, and the completion loss enforces the encoded model to hold semantic information to enable its decoding.
Thus, this regularization is used to enforce the latent space learned by the Siamese network to hold meaningful shape semantic information.

\vspace{-10pt}
\begin{equation}
    \mathcal{L}_{comp} = 
    \sum_{\mathbf{\hat{x}}_i \in \mathbf{\hat{x}}} \min_{\mathbf{\tilde{x}}_j \in \mathbf{\tilde{x}}} \|\mathbf{\hat{x}}_i - \mathbf{\tilde{x}}_j\|_2^2 + 
    \sum_{\mathbf{\tilde{x}}_j \in \mathbf{\tilde{x}}} \min_{\mathbf{\hat{x}}_i \in \mathbf{\hat{x}}} \|\mathbf{\hat{x}}_i - \mathbf{\tilde{x}}_j\|_2^2
    \label{eq:CompletionLoss}
\end{equation}

\subsection{Training}

We pre-train our encoder-decoder network $\Psi(\Phi(\cdot))$ using ShapeNet~\cite{ShapeNet2015} by taking $5997$ samples from the ``car" class.
Our model is fine-tuned by minimizing both tracking and completion losses.
First, we crop and center points lying inside the object's ground truth BB $\{b^t\}_{t \in [1,..,T]}$ for all frames in a given tracklet.
Then, we concatenate the cropped and centered object point clouds to generate an aligned model shape $\mathbf{\hat{x}}$.
Around the ground truth object point cloud at time $t$, we crop a set of $C$ candidate BBs in order to create the candidate shapes $\{\mathbf{x}^t_c\}_{c \in [1,..,C]}$.
The candidate BBs are sampled from a multivariate Gaussian distribution for the three planar degrees of freedom ($t_X$, $t_Y$, $\alpha$) centered around the current object's ground truth BB.

Both the model shape $\mathbf{\hat{x}}$ and the set of candidate shapes $\{\mathbf{x}^t_c\}_{c \in [1,..,C]}$ are encoded into their respective latent representations $\mathbf{\hat{z}}$ and $\{\mathbf{z}^t_c\}_{c \in [1,..,C]}$.
The cosine similarity between the candidates' latent representations $\{\mathbf{z}^t_c\}_{c \in [1,..,C]}$ and the model latent representation $\mathbf{\hat{z}}$ is computed according to \Equation{CosineSimilarity}.
The similarity scores are regressed to their relative Gaussian distance according to \Equation{TrackingLoss}.

Simultaneously, the model shape $\mathbf{\hat{x}}$ is auto-encoded into $\mathbf{\tilde{x}}$ and the Chamfer loss between $\mathbf{\hat{x}}$ and $\mathbf{\tilde{x}}$ is minimized as in \Equation{CompletionLoss}.
Note that we auto-encode the model shape $\mathbf{\hat{x}}$ into itself, instead of encoding the candidate  shapes, as is done for shape completion.
This enforces the latent vector to decode into the most complete car shape we have available, \ie the model shape $\mathbf{\hat{x}}$.

The two losses are minimized jointly as in \Equation{TotalLoss}, with the completion loss being weighted by $\lambda_{comp}$.
We use the Adam optimizer~\cite{kingma2014adam} to train our model with an initial learning rate of $1e^{-4}$, $\beta_1$ of 0.9, and a batch size of 64.
We reduce the learning rate at each plateau for the validation loss using a patience of 
$3$ and a ratio of $0.1$.

% \vspace{-5pt}
\begin{equation}
    \mathcal{L} = \mathcal{L}_{tr} + \lambda_{comp} \mathcal{L}_{comp} 
    \label{eq:TotalLoss}
\end{equation}

\subsection{Testing}
\label{sec:Testing}

Since we are interested in online tracking, 3D tracklets are inferred frame-by-frame.
The shape contained in the tracklet's first BB is used to initialize the model shape $\mathbf{\hat{x}}$.
We track the object by looking over a set of candidate shapes in the frame at time $t$ and comparing them to $\mathbf{\hat{x}}$ using our Siamese network.
The candidate with maximum cosine similarity score is chosen to be the target object for the frame.
The model shape $\mathbf{\hat{x}}$ is then updated by appending to it the chosen candidate shape.
This update step makes the model sensitive to drift, as poorly selected candidates lead to a worse model which subsequently selects worse candidates.
The same problem is encountered in 2D Siamese tracking, commonly solved by not updating the model at all.
However, we show that our model performs better when the model is updated at each frame.

Exhaustively searching for candidates in the three degrees of freedom would incur very high computational cost.
Thus, an approximation of an exhaustive search is leveraged to generate the candidate shapes.
Approximating the exhaustive search allows us to assess the discriminative performance of our Siamese network by assuming the ground truth box will be included as one of the candidates as would be the case with an exhaustive search.
This is a common practice in 2D trackers.
Our exhaustive search is performed by generating candidates using a grid for the three degrees of freedom ($t_X$, $t_Y$, $\alpha$) centered around the current ground truth.
In our experiments, we compare different sampling methods such as Kalman Filters, Particle Filters, and Gaussian Mixture Models, which would be used to provide candidates for our tracker in a more realistic setting.

\section{Experiments}
\label{sec:Experiments}

We use the training set of the KITTI tracking dataset~\cite{geiger2013vision} for our experiments.
It was split as follows: scenes $0$-$16$ were used for training, scenes $17$-$18$ for validation, and scenes $19$-$20$ for testing.
We adapt KITTI  for 3D single object tracking by generating a tracklet for each instance of a car appearing in each of the scenes.
Tracklets are created by concatenating the set of frames in a scene in which a given car instance appears.
For each tracklet, only the first frame includes the ground truth BB.
For our task, we evaluate for Single Object Tracking using the One Pass Evaluation (OPE)~\cite{vot_tpami}.
It defines the \emph{overlap} as the IOU of a BB with its ground truth, and the \emph{error} as the distance between both centers.
The \emph{Success} and the \emph{Precision} metrics are defined using the overlap and error AUC.
For our 3D object tracking purposes, we predict 3D BBs and so we estimate the precision as the AUC for 3D errors from $0$ to $2$m.
We exhaustively generate candidates in a current frame by sampling over a grid of $[-3,3]$m for $t_x$ and $t_y$, and $[-10,10]^o$ for $\alpha$ with a resolution of $1$m and $10^o$, respectively.
The grid is centered around the current ground truth BB to approximate an exhaustive search.
Experiments are run using PyTorch 0.4.1 on a 11GB NVidia GTX1080Ti  GPU.
% Experiments are run using PyTorch 0.4.1 on an NVidia GTX1080Ti with 11GB of memory, in a workstation with up to 64 CPU cores and 256GB of RAM.

\subsection{Ablation Studies}

We present an ablation study of our methodology in \Table{Ablation}, highlighting the importance of the shape completion regularization for the 3D Siamese Tracker.
Results are provided for five different cases: \textbf{(i)} an initialization of our network with random weights, \textbf{(ii)} our network pre-trained on ShapeNet, \textbf{(iii)} our network trained to minimize the completion loss only, \textbf{(iv)} our network trained as a regular Siamese tracker by using only our tracking loss, and \textbf{(v)} our network trained with both the tracking and completion losses.
We observe that training to minimize alone the completion loss or the tracking loss provides better results than pre-training on ShapeNet and a random initialization.
Also, combining both losses enhances the tracker's performance beyond either method isolated.

\begin{table}[htb]
	\centering
	\caption{Ablation study for different losses we are training with. 
	We report the OPE Success/Precision metrics for different losses averaged over 5 runs. 
    Best results shown in bold.}
	\label{tab:Ablation}
	\begin{tabular}{l||c|c} 
 ~~Ablation        & Success      & Precision \\
\midrule
~~(i) Before Training (Random)  & 39.06 & 41.79 \\ \hline
~(ii) Pre-trained on ShapeNet   & 44.54 & 49.38 \\ \hline
(iii) Ours -- Completion only   & 65.36 & 70.62 \\ \hline
(iv) Ours -- Tracking only      & 73.96 & 78.68 \\ \hline
~(v) Ours -- $\lambda_{comp}$@$1e^{-6}$& \bfseries 76.94 & \bfseries 81.38 \\ \hline
    \end{tabular}
\end{table}

\begin{figure}[htb]
    \centering
    \includegraphics[width=0.47\textwidth]{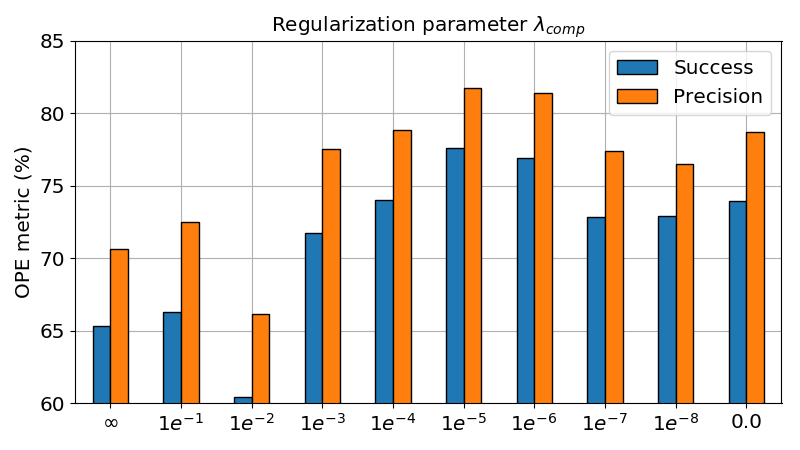}
    \includegraphics[width=0.47\textwidth]{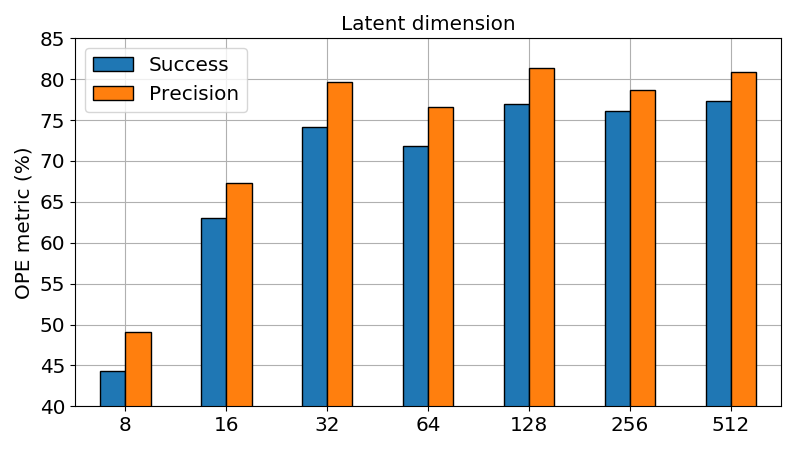}
    \caption{
    Ablation study for different regularization $\lambda_{comp}$ of the shape completion~(\emph{top}) and for the latent representation size~$K$~(\emph{bottom}).
	We report the OPE Success/Precision metrics for different values of $\lambda_{comp}$ and $K$ averaged over 5 runs. 
    }
    \label{fig:Ablations}
\end{figure}

\mysection{Completion Loss.}
\Figure{Ablations} (top) shows detailed results obtained as the regularization parameter $\lambda_{comp}$ is varied.
As less weight is given to the completion loss, the performance moves from the results obtained with only the completion loss to those obtained with only the tracking loss.
The best trade-off is obtained at a point where both losses are in the same order of magnitude.
This occurs with $\lambda_{comp}$ between $1e^{-5}$ and $1e^{-6}$, where we obtain peak performance.

\mysection{Latent representation dimension.}
\Figure{Ablations} (bottom) shows how varying the size of our latent representation $\mathbf{z}$ affects the performances.
It can be observed that a larger latent representation generally performs better.
This is due to the fact that larger latent representations encode more expressive capabilities.
However, this reaches a maximum at a size of around $K=128$ dimensions. 
Larger latent representations require more expensive computations, but the difference is not significant when comparing a latent representation of $32$ dimensions against a $128$-dimensional representation.
Thus, it is best to use the representation which provides the best tracking performance, \ie $K=128$.

% \begin{figure}[htb]
%     \centering
%     \includegraphics[width=0.47\textwidth]{results/3_BottleNeck.png}
%     \caption{Ablation study for the latent representation size $K$. 
% 	We report the OPE Success/Precision metrics for different values of $K$ averaged over 5 runs. 
%     }
%     \label{fig:BottleNeckAblation}
% \end{figure}

% \mysection{Qualitative results.}

\mysection{Reconstruction Performances.}
\Table{Completion} shows shape completion results on the KITTI dataset, using the metrics defined in \cite{Stutz2018VoxelCompletion}.
Our method \textbf{(v)} outperforms the pure completion one \textbf{(iii)} showing that completion also benefits from the different point of view provided during tracking.
However, our decoder is not yet on par with current state-of-the-art.
% as well as the state-of-the-art, showing that our latent space resembles a shape completion space. 

\begin{table}[htb]
	\centering
	\caption{Completion performances on KITTI Tracking.}
	\label{tab:Completion}
	\begin{tabular}{l||c|c|c||c|c} 
        Method            & (iii) & (iv) & (v) & \cite{engelmann2016joint} & \cite{Stutz2018VoxelCompletion}  \\
        \midrule
        Comp. [m]    &  0.188 &  0.690 & \bfseries 0.179 & 0.130 & 0.078  \\ \hline
    \end{tabular}
\end{table}

\begin{figure*}[htb]
    \centering
    \frame{\includegraphics[trim={2cm 1cm 2cm 2cm},clip,width=0.196\textwidth]{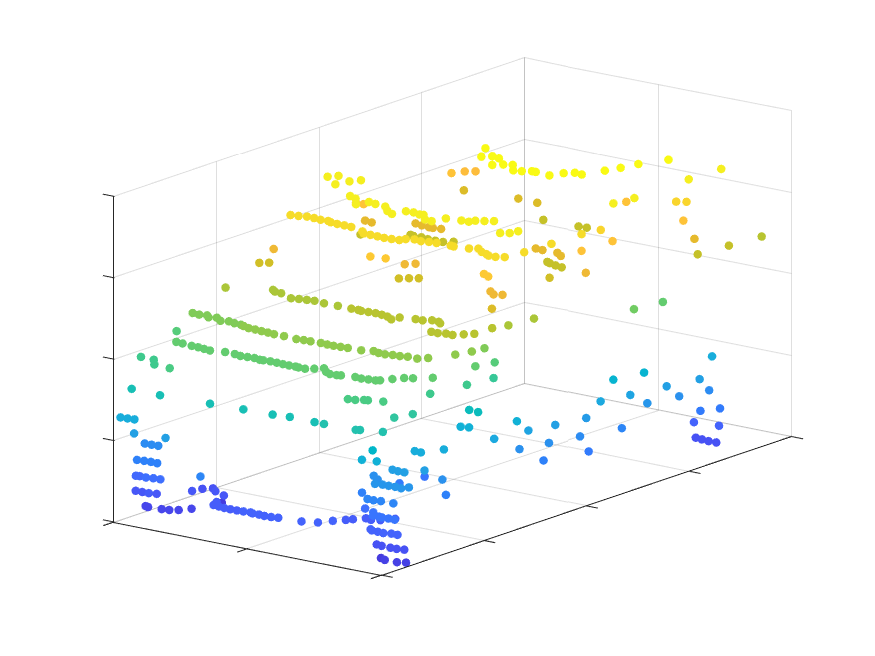}}
    \frame{\includegraphics[trim={2cm 1cm 2cm 2cm},clip,width=0.196\textwidth]{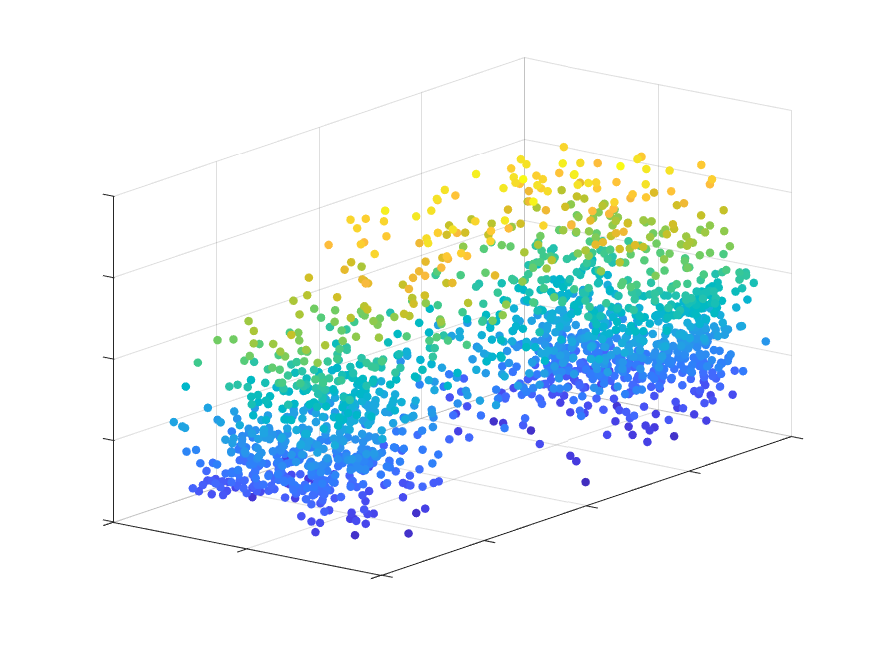}}
    \frame{\includegraphics[trim={2cm 1cm 2cm 2cm},clip,width=0.196\textwidth]{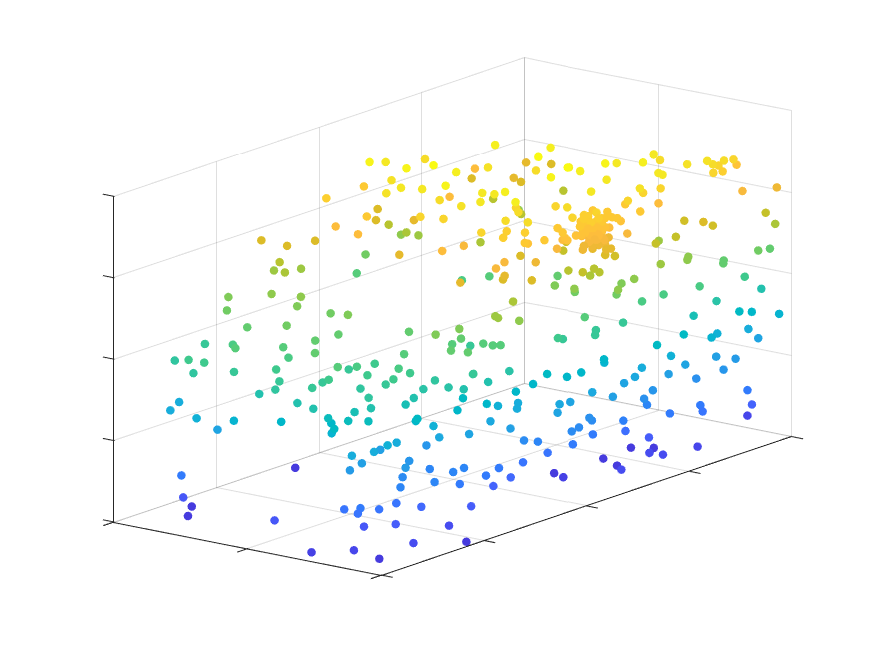}}
    \frame{\includegraphics[trim={2cm 1cm 2cm 2cm},clip,width=0.196\textwidth]{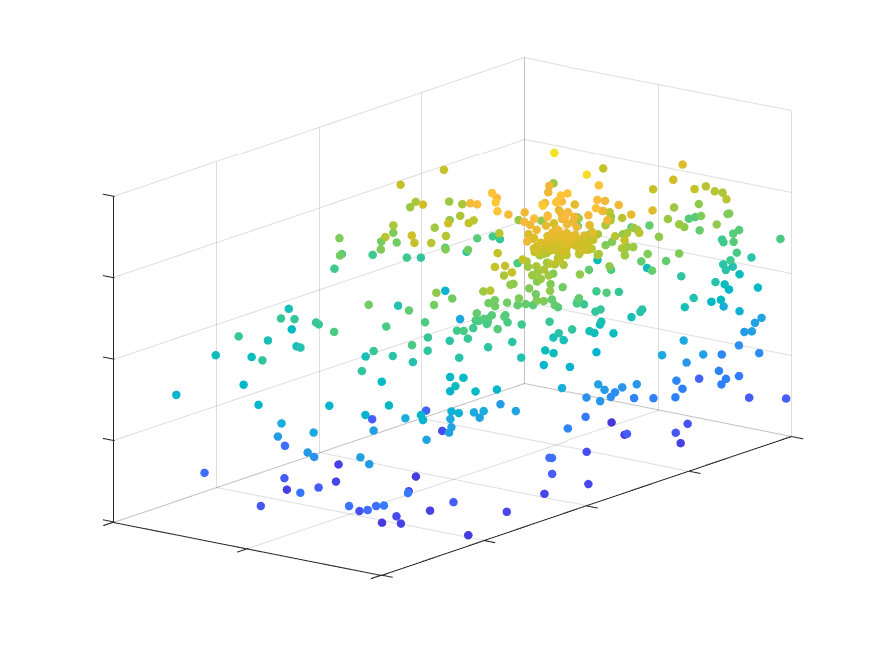}}
    \frame{\includegraphics[trim={2cm 1cm 2cm 2cm},clip,width=0.196\textwidth]{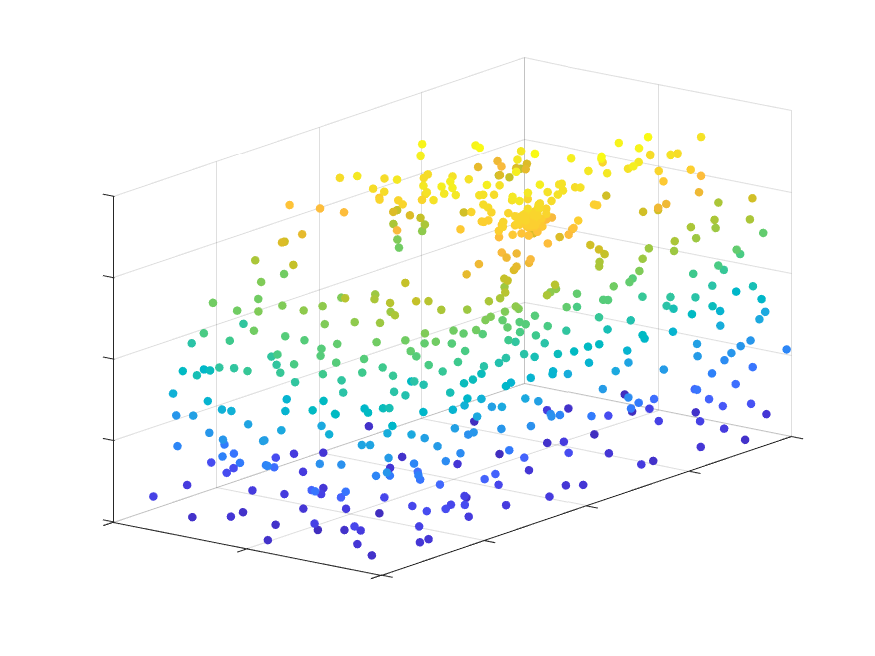}}
    \caption{Example of model completion (from left to right): 
    \textbf{(i)} Candidate point cloud, 
    \textbf{(ii)} Decoded candidate point cloud when it is pre-trained with ShapeNet, 
    \textbf{(iii)} Decoded candidate point cloud when it is trained with completion loss only ($\lambda_{comp} = \infty$),
    \textbf{(iv)} Decoded candidate point cloud when it is trained with tracking loss only ($\lambda_{comp} = 0$) (the decoder trained for completion is used for fair comparison),
    \textbf{(v)} Decoded candidate point cloud when it is trained with both tracking and completion losses ($\lambda_{comp} = 1e^{-6}$).
    }
    \label{fig:ModelCompletion}
\end{figure*}

\mysection{Qualitative results.}
\Figure{ModelCompletion} shows qualitative results regarding decoded shapes $\Phi(\mathbf{x})$.
We can observe that training for tracking only results in a decoded point cloud containing a large amount of noise.
Already, the model pre-trained on ShapeNet provides a reconstruction which resembles a general car but not the specific candidate car.
Training for shape completion only provides a shape reconstruction which is a more complete version of the original candidate shape.
Regularizing tracking with shape completion by using $\lambda_{comp}=1e^{-6}$ provides a reconstruction similar to that using shape completion only.
However, the model trained for shape completion only follows the candidate shape more closely.
A regularized loss is able to improve tracking results while conserving enough class information as to reconstruct the encoded shape from its latent vector.

\begin{figure}[htb]
    \centering
    \includegraphics[trim={4.5cm 8cm 4cm 9.5cm},clip,width=0.47\textwidth]{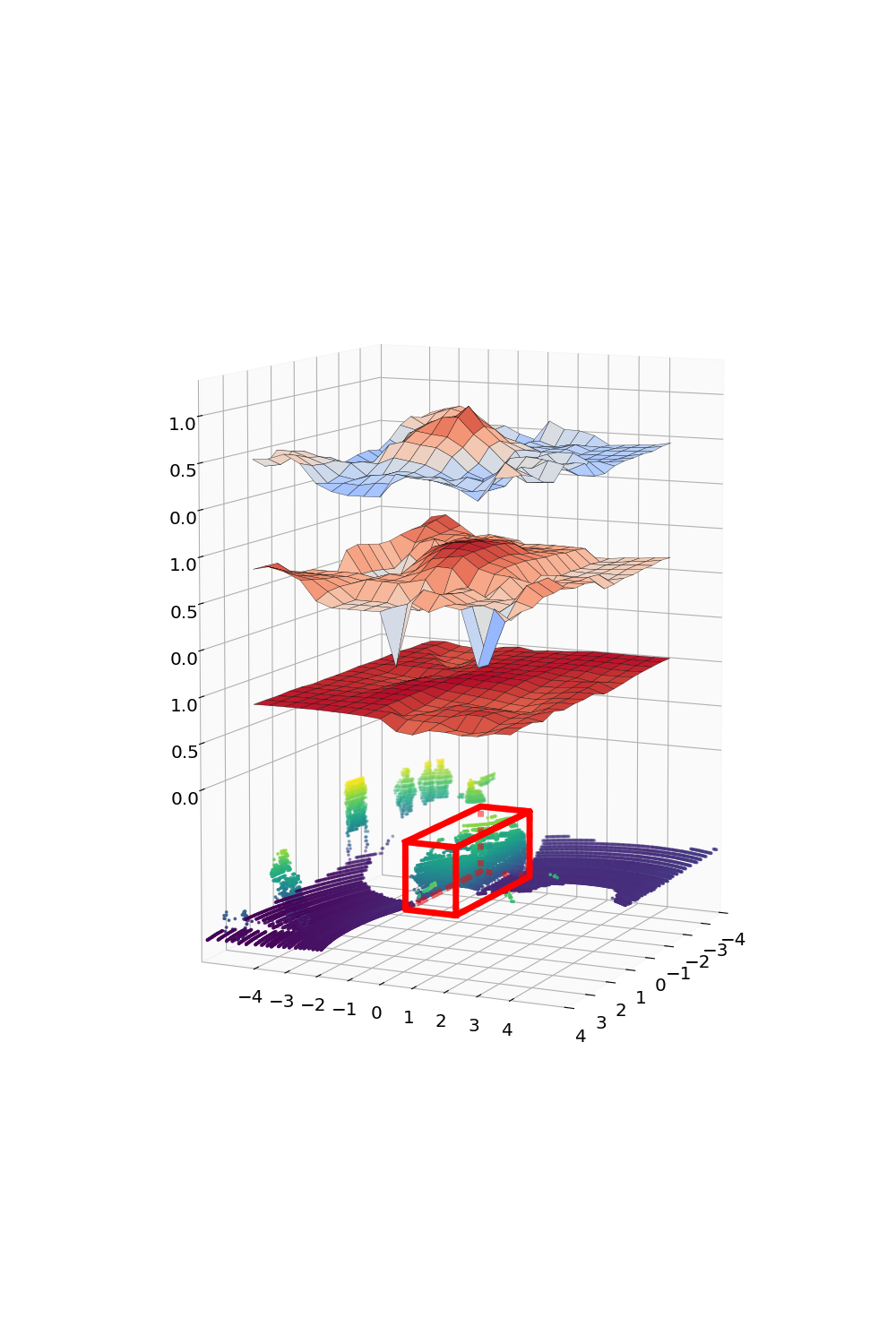}
    \caption{Heatmap of model cosine similarity scores on an exhaustive search space grid:
    From bottom to top: 
    \textbf{(i)} activation using random weights model,
    \textbf{(ii)} activation on pre-trained model (ShapeNet),
    \textbf{(iii)} our model.
    }
    \label{fig:HeatMaps}
\end{figure}

\Figure{HeatMaps} illustrates the activations obtained from the cosine similarity for a set of samples obtained around an exhaustive search.
We observe that a randomly initialized model generates high scores everywhere, hence providing a bad discrimination.
A model pre-trained on ShapeNet is able to better discriminate the shape to track than random initialization, but is still distracted by the environment, confusing other shapes for the car.
Our model is able to discriminate fairly well between the ground truth car and the surrounding areas; there are high activations only in the vicinity of the ground truth box.
Note that the ideal shape we expect to obtain for the activations is a Gaussian centered at the ground truth BB, as regressing to in training.

\subsection{Model Fusion and Shape Aggregation}

We construct and update a model $\mathbf{\hat{x}}$ for the target object's shape as we track it.
By default, the model is maintained as a point cloud.
Our update step for the model after iteration $t$ consists of concatenating the points of the tracked shape in frame $\mathbf{x}^t$ with the current model.
An alternative is to maintain a model by averaging the latent representations.
We investigate the effects of fusing either point clouds or latent representations as well as the effects of different types of aggregation in time for both representations.
We test the different shape fusions and aggregations in our method.
We report the main results in \Table{Fusion}.

\begin{table}[htb]
	\centering
	\caption{OPE Success/Precision for different Data Fusion and Model Aggregation.
    All results are averaged over 5 runs. 
    Best representation aggregation shown in bold.}
	\label{tab:Fusion}
	\begin{tabular}{l||c|c} 
Fusion              & Early Fusion & Late Fusion \\
Data Representation     & PC         & Latent \\
\midrule
First shape only        &54.6 / 64.2    &54.6 / 64.1 \\ \hline
Previous shape only     &64.5 / 69.7    &64.4 / 69.6 \\ \hline
First and prev. shapes  &75.4 / \textbf{82.7}    &69.1 / \textbf{78.1} \\ \hline
All previous shapes     &\textbf{76.9} / 81.4    &63.9 / 73.2 \\ \hline
Median Pooling          &-- / --    &59.7 / 67.6 \\ \hline
Max Pooling             &-- / --    &\textbf{71.5} / 75.6 \\ \hline
    \end{tabular}
\end{table}

\mysection{Early/Late Fusion.}
We update the model by either concatenating the shape point clouds $\mathbf{x}^t$ (\emph{Early Fusion}) or aggregating the latent shape representations $\mathbf{z}^t$ (\emph{Late Fusion}).
Early Fusion requires a larger amount of memory to store the model shape.
Late Fusion allows for a more memory-efficient representation for point clouds, since we only need to keep a latent vector to represent a whole shape.
It is also more computationally efficient since the model is not encoded several times during testing.

\mysection{Shape Aggregation.}
We investigate different types of shape aggregations.
In particular, we try using the shape in the first frame only, the previous shape only, an aggregation of the the first and previous shapes, and an aggregation of all the previous shapes.
We also investigate aggregating the latent representations by either computing the average, the median, or the max of the vectors across time.

\mysection{Analysis.}
As shown in \Table{Fusion}, concatenating point clouds (Early Fusion) performs generally better than fusing the latent vectors (Late Fusion).
This is mainly due to our completion loss designed to handle arbitrary shapes sampled at $2048$ points.
We do not include any loss that would train our network to aggregate latent vectors.
As a result, late fusion does not perform as well as concatenating point clouds.

Using an aggregation of only the first or the previous frame does not perform well.
In particular, the number of points belonging to the object in question in a single frame can be significantly small, which impedes a proper shape representation.
Should this happen in the first frame, it will imply a bad initial representation.
A low point count when tracking using the previous frame will induce drifting.

Fusing the first and previous frames performs surprisingly well and provides the best precision.
We believe that the two distant representations complement each other, particularly by limiting the amount of translational drift in the first frames given an initial bad representation.
The shapes in the first frames typically contain a limited number of points, since they are sensed from a large distance.
They provide a very incomplete shape information, but are still helpful to localize roughly its position although not its orientation.
The full model will inevitably drift while the fusion of first and previous frame avoids initial drifts to a certain extent, thus having an improved precision.

For the latent representation, median pooling is less effective than average pooling, but max pooling provides the best performance.
We argue that it interacts well with the max pooling layer at the end of our encoder network.
By consecutively max pooling over the shape's point features (last layer of our encoder) and over all the previous latent vectors, we actually pool over all the shape's point feature in a tracklet, which provide a more global model latent representation.
Still, this is not as effective as Early Fusion.

\subsection{Search Space}

Defining an efficient search space is extremely difficult in 3D due to the continuity and cubic nature of 3D space.
Thus, an exhaustive search becomes infeasible when a very fine search space is required.
To overcome this limitation, we use a Kalman Filter, Particle Filter, and Gaussian Mixture Model to generate candidates.
We apply our network using more realistic search spaces, which do not use the ground truth BB, as opposed to the exhaustive search approximation.
We argue that our model has good discriminating capabilities, but is limited by the quality of proposed candidates.
To support our claim, we report the results obtained by scoring the candidates using their distance to the ground truth object BB -- the best possible similarity metric -- along with the results obtained using our best model with both early fusion and late fusion.
Results are shown in \Table{SearchSpace}.
It can be observed that our model reaches performances similar to those obtained by selecting the candidate closest to the ground truth, which emphasize the effectiveness of our similarity metric for discrimination.

\begin{table}[htb]
	\centering
	\caption{OPE Success and Precision for different Search Space.
    All results are averaged over 5 runs.}
	\label{tab:SearchSpace}
	\begin{tabular}{l||c|c|c}
Fusion          & Early         & Late          & Closest \\
Data Repres.    & PC            &Latent         &Space \\ \midrule
Kalman Filter   & \textbf{41.3} / \textbf{57.9}   & \textbf{37.4} / \textbf{52.1}   & \textbf{43.7} / \textbf{58.3} \\ \hline
Particle Filter & 34.2 / 46.4   & 33.3 / 44.9   & 38.4 / 49.5 \\ \hline
GMM(k= 25)      & 35.6 / 49.1   & 34.0 / 46.1   & 37.9 / 49.3  \\ \hline
	\end{tabular}
\end{table}

\subsection{Comparison with Baselines}
To compare our method for 3D tracking, we create two baselines due to the absence of 3D tracking methods for this specific task.
We take as baselines a \sota 3D detection method as well as a 2D tracker.
The results from these baselines are reported along with our best model using exhaustive search and our best model using a Kalman filter in \Table{ComparisonTrackingBaseline}.
Evaluation metrics are reported using both the 3D IOU on 3D BBs and the 2D BEV IOU on BEV BBs.
% The 2D BEV IOU metrics are reported in the same fashion as the 3D IOU, but use the 2D BEV BBs instead of the 3D BBs.

\begin{table}[htb]
    \centering
    \caption{Baseline comparison using the 3D OPE (3D BB) and the 2D OPE on BEV frames.}
    \label{tab:ComparisonTrackingBaseline}
    \begin{tabular}{l||c|c} 
        Test                 & OPE$_{3D}$   & OPE$_{2D}$ \\ 
        \midrule
        STAPLE$_{CA}$           & -- / --        & 31.60 / 29.30     \\ \hline
        AVOD Tracking        & 63.16 / 69.74  & 67.46 / 69.74   \\   \hline
        Ours - Kalman Filter & 40.09 / 56.17  & 48.89 / 60.13              \\ \hline 
        Ours - Exhaustive    & \textbf{76.94 / 81.38} &  \textbf{76.86 / 81.37}     \\ \hline 
    \end{tabular}
\end{table}

\mysection{3D Detection.}
For the 3D detection baseline, we pair the AVOD-FPN~\cite{ku2017joint} detector with an online matching algorithm.
AVOD-FPN utilizes both LIDAR point clouds and RGB images to obtain 3D detections.
We use the detection for every frame in our tracklets and preform tracking-by-detection by matching objects frame-by-frame.
The object in frame $t$ is selected as the BB with the highest overlap with the BB tracked in frame $t-1$.

\mysection{2D Tracker.}
We compare against the popularized 2D STAPLE$_{CA}$ tracker \cite{bertinetto2016staple,mueller2017context}, when applied to BEV data.
BEV images are extracted from point clouds in our tracklets by projecting points into the ground plane.
The resulting 2D tracklets are then fed to the STAPLE$_{CA}$ tracker.
This method provides a LIDAR-only tracker as a fair baseline for our method, which also only relies on LIDAR input.

\mysection{Analysis.}
\Table{ComparisonTrackingBaseline} shows the comparative results with tracking baselines.
Our exhaustive model performs better than both baselines, while the model using a Kalman filter is able to outperform the 2D Tracker.

\section{Discussions}
\label{sec:Discussion}

\mysection{Training on complete models.}
In our experiments, we auto-encode a complete model shape obtained by concatenating all the point clouds in a tracklet.
We then enforce candidate shapes belonging to the same object as our model to encode into a vector with a high cosine similarity between itself and the latent representation of our model.
An alternative would be to enforce partial shapes belonging to an object at different times to be similar to each other and partial shapes not belonging to the object to be dissimilar from those belonging to the object.
In particular, we attempted to provide the object at time $t$ as a target for our Siamese network in place of the full model.
However, better results were obtained by using a full model as the target.

A natural extension to training using objects from the same time $t$ is to concatenate different combinations of shapes from the same tracklet at different times.
This augmentation is possible since we train our network to complete shape \ie to be invariant to occlusions from different views.
This is an intermediate step between training using a single frame as a model and using the whole tracklet to create a model shape for the Siamese network.
However, this augmentation increased training time exponentially and did not provide further improvements to our tracking results.
There are not enough points in every frame to learn the proper shape of the car without auto-encoding a full model.

\mysection{Ground included in the car model.}
The model is pre-trained on Shape net, which has complete shapes without noisy points such as a road.
In our test, we scaled the BBs by a factor of 1.25 since the original BBs are too tight around the cars and part of the border of cars lie outside their BBs.
Such consideration account for $10$\% of the performances.
For this reason, it is possible to see the road in \Figure{ModelCompletion}, but we believe that including the road does not negatively affect the shape representation.
We also considered a fixed offset of $0.5$m which proved to be less effective.

% \mysection{Gaussian distance vs IoU.}
% We have tried to regress the 3D IoU between candidate and model BBs instead of regressing a Gaussian distance.
% However, using the IoU as a target for regression led to worse performances.

\mysection{Robustness on Occlusion.}
With the tracking loss only, our method (iv) performs a Success/Precision of $76.9/80.1$ and $72.8/77.4$ for fully visible and occluded samples in KITTI, respectively.
Adding shape completion, our method (v) reaches $79.9/83.2$ and $74.6/80.5$, showing improvement in both cases.

\mysection{Robustness on Dynamic Scenes.}
We computed the dynamics of each vehicle to track in KITTI, and report an average distance $d=0.742m$ between consecutive frames, in agreement with Figures~10\&11 of \cite{geiger2013vision}.
We split our samples into a static ($d<0.7m$) and a dynamic ($d>0.7m$) set from which we report fairly similar Success/Precision metrics of $76.4/80.5$ and $76.7/83.2$ respectively.

\mysection{Symmetry.}
Most cars are visible only from one side.
We attempted leveraging a prior knowledge of car symmetry in order to complete furthermore the shape of the cars.
However, this method did not prove to be effective, in particular because the BBs are not well centered and introduce more noise into our model.

\mysection{Gaussian Sampling.}
We generate candidate offsets during training by sampling from a multivariate Gaussian distribution, in contrast with sampling offsets using a fixed grid.
Sampling offsets randomly improves performances, since the network is able to learn from a variety of target scores.
Fixing an offset grid provides only a discrete number of target scores used for the tracking loss.
Lacking variety in training induces worse performances during testing.

\mysection{Timing.}
Our model takes on average 1.8ms to evaluate 147 candidates.
We do not account for the time spent generating and preparing the candidates and model point clouds for evaluation.
This allows us during deployment to increase the number of candidates as much as allowed by the GPU, while still being able to process point clouds in real-time.

\section{Conclusion}

In this paper, we propose, to the best of our knowledge, the first 3D Siamese tracker applied to point clouds rather than images.
We leverage an efficient encoding able to embed meaningful semantic priors thanks to a shape completion regularization.
We show that regularizing our network with semantic information results in better discrimination and tracking performances.
Also, we provide insights on model building such as early/late fusion and shape aggregation in frames. 
We compare against baselines in 3D and 2D BEV, showing that our discriminator is able to outperform baselines by using exhaustive search settings.
As a result, we propose a purely 3D alternative for tracking cars in urban environments, and show that geometric-oriented approaches are capable of attaining good performances.

Future works will also include improving both the similarity metric and the model update, by including a proposal loss similar to that used in region proposal networks and a smarter model point cloud selection based on the quality of point clouds.
Further works will include an extension to Multiple Object Tracking and 3D Object Detection, by leveraging the similarity metric based on our 3D Siamese network.
Alternatively, 3D Siamese tracking could be adapted to different classes of objects, articulated shape representation and 2D object tracking.

\mysection{Acknowledgments:}
This publication is based upon work supported by the King Abdullah University of Science and Technology (KAUST) Office of Sponsored Research (OSR)
under Award No. RGC/3/3570-01-01.
% under Award No. OSR-CRG2017-3405
% RGC/3/3570-01-01

{\small
\bibliographystyle{ieee}
\bibliography{biblio}
}

\end{document}